\documentclass[10pt,twocolumn,letterpaper]{article}
\usepackage{cvpr}
\usepackage{times}
\usepackage{epsfig}
\usepackage{graphicx}
\usepackage{amsmath}
\usepackage{amssymb}
\usepackage{float}
\usepackage{multirow}
\usepackage{tikz}
\usepackage{gensymb}
\newcommand*{\circled}[1]{\lower.7ex\hbox{\tikz\draw (0pt, 0pt)%
    circle (.5em) node {\makebox[1em][c]{\small #1}};}}

\usepackage[breaklinks=true,bookmarks=false]{hyperref}

\cvprfinalcopy % *** Uncomment this line for the final submission

\def\cvprPaperID{8738} % *** Enter the CVPR Paper ID here

\begin{document}

%%%%%%%%% TITLE
\title{Where, What, Whether: Multi-modal Learning Meets Pedestrian Detection}

\author{Yan Luo,
Chongyang Zhang,
Muming Zhao,
Hao Zhou,
Jun Sun\\
%$^{1}$School of Electronic Information and Electrical Engineering,\\
%Shanghai Jiao Tong University, Shanghai 200240, China\\
%$^{2}$MoE Key Lab of Artificial Intelligence, AI Institute, Shanghai Jiao Tong %University, Shanghai 200240, China.\\
%{$^*$Corresponding author: Chongyang Zhang, sunny\_zhang@sjtu.edu.cn}
}
\twocolumn[{%
\maketitle
\begin{figure}[H]
\hsize=\textwidth
\centering
\includegraphics[width=15.5cm]{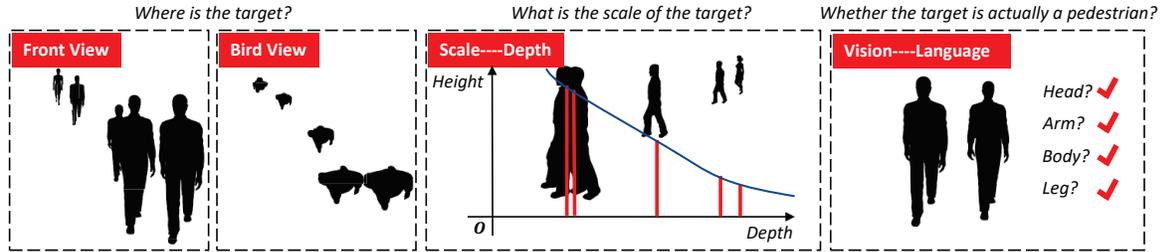}
\caption{W$^3$Net decouples the pedestrian detection into \textbf{\textit{W}}here, \textbf{\textit{W}}hat and \textbf{\textit{W}}hether problem as illustrated above from left to right, which enables us to generate robust representations against occlusion and scale variation.}
\label{fig:motivation}
\end{figure}
}]
\pagestyle{empty}
\thispagestyle{empty}
\begin{abstract}
Pedestrian detection benefits greatly from deep convolutional neural networks (CNNs). However, it is inherently hard for CNNs to handle situations in the presence of occlusion and scale variation. In this paper, we propose W$^3$Net, which attempts to address above challenges by decomposing the pedestrian detection task into \textbf{\textit{W}}here, \textbf{\textit{W}}hat and \textbf{\textit{W}}hether problem directing against pedestrian localization, scale prediction and classification correspondingly. Specifically, for a pedestrian instance, we formulate its feature by three steps.
i) We generate a bird view map, which is naturally free from occlusion issues, and scan all points on it to look for suitable locations for each pedestrian instance.
ii) Instead of utilizing pre-fixed anchors, we model the interdependency between depth and scale aiming at generating depth-guided scales at different locations for better matching instances of different sizes.
iii) We learn a latent vector shared by both visual and corpus space, by which false positives with similar vertical structure but lacking human partial features would be filtered out.
We achieve state-of-the-art results on widely used datasets (Citypersons and Caltech). In particular. when evaluating on heavy occlusion subset, our results reduce MR$^{-2}$ from 49.3$\%$ to 18.7$\%$ on Citypersons, and from 45.18$\%$ to 28.33$\%$ on Caltech.

\end{abstract}

\section{Introduction}
Pedestrian detection is a fundamental topic in computer vision. Generally speaking, the design of pedestrian detector is deeply influenced by the development of object detection, which is used to tell \textit{\textbf{where}} the object is, and \textit{\textbf{how}} big it is~\cite{liu2018CSP}. Most modern anchor-based detectors~\cite{Cheng2017Pedestrian}~\cite{Zhang2018Attention}~\cite{Zhang2018Occlusion} get stuck in a paradigm, namely, expertise-based techniques to generate a series of anchors and then identify whether it is a pedestrian or not. It is undeniable that this method has a extensive impact and is widely used as a powerful baseline, but a large number of redundant and low-quality proposals it introduces also limit the accuracy and speed. In contrast, our work falls into the anchor-free fashion, which takes multi-modal data as input to predict \textit{\textbf{where}} the target is, \textit{\textbf{what}} the scale is, and \textit{\textbf{whether}} the target is actually a pedestrian, instead of utilizing pre-fixed anchors.

\textbf{\textit{Where }}- Object detection actually starts from a basic problem: where is the target? In previous successful practices, such as Faster RCNN~\cite{Ren2017Faster} or SSD ~\cite{Liu2015SSD}, this process is mostly determined by a set of pre-defined anchors, based on a implicit dependency that objects are evenly distributed on the image, which is much similar to exhaustion. The recently popular anchor-free methods shake of the yoke~(pre-defined anchors) and detect objects directly from an image, which makes the detection a more natural way. One of the typical and effective anchor-free practice in pedestrian detection is CSP~\cite{liu2018CSP}, which facilitated the \textit{Where} problem as a straightforward \textit{center prediction} task. Our work is conceptually similar to CSP, which is also in the scope of anchor-free, but differs significantly in insights. As the author states in CSP~\cite{liu2018CSP} (Section 4.4), the pedestrian center is actually a fuzzy point of semantic information, in which the variation of pedestrian wearing or orientation will have a negative effect on it. In addition, we discover that the pedestrian center is vulnerable to occlusion, that is to say, in some occluded scenes, the pedestrian center is invisible. The above challenges motivate us to find a unified and robust representation for locations of pedestrians. In this paper, we attempt to attribute occlusion issues to the interlocking problem caused by the single image view (front view). If the 2D image is switched to the bird view map, occlusion would be greatly alleviated. As illustrated in Figure~\ref{fig:motivation} (left), even if a pedestrian has occlusion in the front view, the bird view map still performs free from occlusion issues.

\textbf{\textit{What }}- Another long-standing problem exists in object detection as well as pedestrian detection is what the scale of the target is. For the anchor-based methods, obviously, they rely more on a set of pre-defined scales and aspect ratios. Or, the anchor-free method such as CSP, stacks with convolutions to predict pedestrian scales and set aspect ratio as a uniformed value of 0.41~\cite{liu2018CSP} (Section 3.3). Despite these pipelines have been shown effective in several benchmarks, the exhaustive proposal sizes in anchor-based methods appear redundant, and the uniform scale setting, such as 0.41 in CSP~\cite{liu2018CSP}, seems to be inflexible. In this paper, the proposed W$^3$Net is motivated by the discovery that the scale distribution of pedestrians in the 2D image is not out of order, on the contrary, is closely related to its geometry. As is shown in Figure~\ref{fig:motivation}(middle), intra-image pedestrian scale variations vary along with the estimated depth (the distance from the camera in real world), which means the nearer instances should have lager scales compared to farther one. Following this intuition, we model the interdependency between depth and scale aiming at generating proper scales that flexible and accurate at different locations.

\textbf{\textit{Whether }}- Neither anchor-based nor anchor-free methods can stay out of the problem: whether the bounding box actually filter the pedestrian. Due to the existence of occlusion issues and the diversity of occlusion patterns, instance features used for the downstream classification show an obvious difference, and consequently make the "\textit{Whether}" problem difficult. Some efforts have been made to handle this matter. Part-based methods~\cite{Zhang2018Attention}~\cite{Zhang2018Occlusion} tends to employ a weakly-supervised manner to perceive the visible body parts, while the dual branch method~\cite{Zhou2018Bi} proposes two sub-networks, one for full body estimation and the other for visible part. These methods, in the final analysis, treat the occluded and non-occluded pedestrians in a split way, which suffers uncertainty of weakly-supervised part labels and imbalance between the number of occluded and non-occluded samples. In contrast, we discover that pedestrians with obvious attributes can be represented by corpus, such as head, arm, body and leg, which offers a possibility to re-encode both occluded and non-occluded instances into a unified corpus space, and thus benefits robust feature generation against "\textit{\textbf{Whether}}" problem.

W$^3$Net is evaluated with challenging settings on Citypersons~\cite{Zhang2017CityPersons} and Caltech~\cite{Dollar2009Pedestrian} pedestrian dataset, and still achieves state-of-the-art performance. In particular, when evaluating heavy occlusion subset on Citypersons, our result reduces MR$^{-2}$ by a factor of 2 (our model produces 18.7\% while prior arts range from 49\%-56\%).

\section{Related Work}
 With multi-modal learning, the proposed W$^3$Net confronts challenges (occlusion and scale variation), and decouples the task into \textbf{\textit{W}}here, \textbf{\textit{W}}hat and \textbf{\textit{W}}hether problem directing against pedestrian localization, scale prediction and classification. Therefore, we review recent work on pedestrian detection with or without multi-modal data and compare ours with previous state-of-the-art methods.

Most prevalent pedestrian detectors are on the basis of the framework in general object detection such as Faster RCNN~\cite{Ren2017Faster} and SSD~\cite{Liu2015SSD}, and leverage pedestrian-oriented traits to tackle challenges such as occlusion and scale variation in the pedestrian detection task. Part-based methods, for example DeepParts~\cite{felzenszwalb2010dpm}, FasterRCNN+ATT~\cite{Zhang2018Attention} or OR-CNN~\cite{Zhang2018Occlusion}, fully exploit pedestrian part information especially visible body parts to assist robust feature embeddings. With the careful discovery that pedestrians usually perform up-right posture, TLL~\cite{TLL2018} escapes from pre-defined anchors and proposes the line localization, which greatly promotes the development of pedestrian detection.

\begin{figure*}
\begin{center}
\includegraphics[width=17cm]{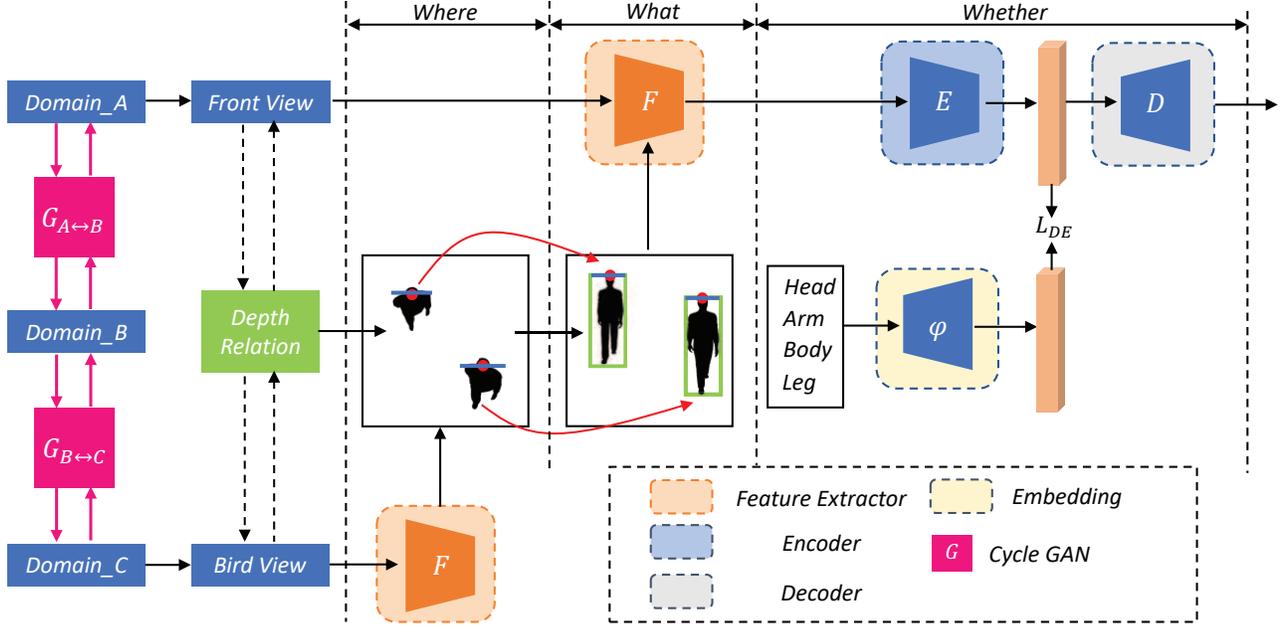}
\end{center}
   \caption{Overview of the proposed W$^3$Net. "\textbf{\textit{Where}}": Predict the locations of the head (red dots) and width (blue lines) on \textbf{\textit{bird view map}}. "\textbf{\textit{What}}": Depth-guided proposal alignment from \textbf{\textit{bird view map}} to \textbf{\textit{front view map}}. "\textbf{\textit{Whether}}": Features extracted from \textbf{\textit{front view map}} are re-encoded to a unified corpus space. The "\textbf{\textit{Whether}}" branch is optimized by $L_{DE}$ to learn a latent space shared by both visual features and attribute embeddings. The overall output is used for the downstream pedestrian localization and classification.}
\label{fig:framework}
\end{figure*}

Recently, some researchers have turned attention to multi-modal learning, which offers the possibility of capturing correspondences between multiple modalities and gains an in-depth understanding of natural phenomena. SDS-RCNN ~\cite{SDSRCNN2017} proposes a segmentation infusion network to enable joint supervision on semantic segmentation and pedestrian detection. F-DNN+SS ~\cite{Du2017Fused} uses a derivation of the Faster RCNN framework, and further incorporates pixel-wise semantic segmentation in a post-processing manner to suppress background proposals. F-DNN2+SS ~\cite{Du2018Fused} employs an ensemble learning approach, and semantic segmentation network to adjust the confidence in the detector proposals.

Motivated by above, the proposed W$^3$Net takes a further step in three-folds:

\emph{1)} We first attempt to explore a new possibility that pedestrians can be effectively described in the bird view map, which bypasses the limitation of 2D (front) images and benefits the generation of robust features against various occlusion patterns.

\emph{2)} We first attempt to model interdependency between depth and scale for pedestrian detection, which predicts flexible and proper proposals other than predefined ones.

\emph{3)} We first attempt to propose a method to embed pedestrian features in corpus space, which aligns both occluded and non-occluded instances.

\section{Methodology}
W$^3$Net falls into the anchor-free fashion, which decouples the pedestrian detection task into three sub-problems. With the input of multi-modal data, including bird view map, depth and corpus information, the three problems "\textbf{\textit{Where}}", "\textbf{\textit{What}}" and "\textbf{\textit{Whether}}" target at pedestrian localization, scale prediction and classification to benefit robust feature generation against occlusion and scale variation.

The pipeline is shown in Figure~\ref{fig:framework}. Specially, taking the front view image $I_{F}$ (Domain\_A) as input, the network first generates the corresponding bird view map $I_{B}$ (Domain\_C) through two Cycle GAN~\cite{CycleGAN2017}. The $I_B$ is then processed by the forward feature extractor to predict probability of pedestrians at each location and the corresponding width, denoted as "\textbf{\textit{Where}}" branch. The connecting "\textit{\textbf{What}}" branch, supported by depth relation, aligns results from "\textbf{\textit{Where}}" branch and generating reasonable proposals for the following "\textbf{\textit{Whether}}" branch without any \textit{conv} or \textit{fc} effort. Proposal features extracted by feature extractor are subsequently fed into the encoder-decoder, in which the latent space is occupied by corpus and promotes both occluded or non-occluded instances to be re-encoded by common attributes of pedestrians. Overall, the three "\textbf{\textit{W}}" branches link together, complement each other, and constitute the proposed W$^3$Net.

\subsection{Bird view: Where the target is}
\label{section:bird}
 Compared with front view, bird view owns the unique advantages: \emph{1)} The bird view map is naturally free from occlusion issues. We discover that occlusion, whether it is intra-class or inter-class~\cite{Wang2017Repulsion}, could be mostly attributed to the single detection view (front view). \emph{2)} Pedestrians are more likely to be condensed into a unified feature representation through bird view, namely 'head'. Previous anchor-free method, such as CSP which suffers difficulty to decide an 'exact' center point during training, the proposed 'head' prediction on bird view map appends more specific and accurate classification targets for the "\textbf{\textit{Where}}" branch.

 However, how to obtain the bird view map still remains to be solved, which we treat as an image-to-image generation process in this paper. Due to the lack of real bird view map in most of existing pedestrian datasets, such as Citypersons~\cite{Zhang2017CityPersons} and Caltech~\cite{Dollar2009Pedestrian}, we attempt to introduce synthetic data, which directly captured from 3D games to train a bird view generation model. As the previous practice in ~\cite{abarghouei18monocular}, synthetic data such as ~\cite{Param2017ObjSyn}~\cite{Tuan2017Sythetic} faces a huge challenge: the performance would be limited if the model trained on synthetic data is directly used on real-world data. Inspired by above, we design our front-to-bird view generation network by two steps. The first includes a CycleGAN~\cite{CycleGAN2017}~\cite{isola2017image}, denoted as $G_{A\rightarrow B}$ in Figure~\ref{fig:framework}, to transfer real-world data (Domain\_A) to the synthetic images (Domain\_B). The following step introduces the other CycleGAN, denoted as $G_{B\rightarrow C}$, to train a bird view generation model over synthetic data captured from the game of GTA5, which consists of a total 50,000 pairs of images including front view and bird view from virtual car. Based on the well-trained Generative Adversarial Network (GAN)~\cite{Goodfellow2014Generative} $G_{A\rightarrow B}$ and $G_{B\rightarrow C}$, the input front view map $I_F$ could be transformed to the bird view map $I_B$.

 Upon $I_B$, a detection head, which is composed of one $3\times3$ \textit{conv} and two $1\times1$ \textit{conv} layers, is attached behind the feature extractor to predict probability of pedestrians at each location and the corresponding width. With its inherent superiority on occlusion, the bird view map can bypass the limitation of front view images, and thus benefit the generation of robust features. However, it is not well performed yet. On the one hand, bird view map pays for the loss of pedestrian height information, which brings about unsatisfied height prediction results on the single bird view map, and thus derives the following "\textbf{\textit{What}}" branch to estimate flexible and accurate scales for different instances. On the other hand, bird view map suffers from \textit{false positives} caused by suspected objects, such as the top of street lamps \textit{vs.} pedestrian 'head', thus the introduced "\textbf{\textit{Whether}}" branch is directed against this issue.

\subsection{Depth: What the scale is}
As is known to us, the bounding box has four-degree-of-freedom, that is $\left \{ x,y,w,h \right \}$. During the process of proposal generating, $\left \{ x,y \right \}$ are already known, which are evenly distributed on the feature map, and the corresponding $w$ has been predicted on the bird view. Therefore, the "\textbf{\textit{What}}" branch only includes two tasks: \emph{1)} What the width is on the front view map. \emph{2)} What the corresponding height is.

Subsequently, to address above tasks, we start from a more general situation with the following definition.

\textit{\textbf{Definition:}} \textit{We define $v_{0,0}$ as the front view when the image $I_F$ is captured at the front horizontally, and $v_{H, \theta}$ as the view when the image $I_B$ is captured with the depression angle $\theta$ at the height of $H$. Specially, if $\theta$ is $90\degree$, the view $v_{H,90\degree}$ looks towards perpendicular to horizontal.}

On the above basis, we could build up a set of geometric relations. We suppose the real world 3D coordinates of the pedestrian \textit{P} under the view of $v_{0,0}$ is $(X,Y,Z)$, in which $Z$ could be also considered as the distance from the camera, that is \textbf{depth},while the corresponding 2D coordinates on the image $I_F$ is $(u,v)$, in which given the camera intrinsic matrix $K$, the projection from $(u,v)\rightarrow(X,Y,Z)$ can be formulated as the following:
\begin{equation}
[X,Y,Z]^{T}=ZK^{-1}[u,v,1]^{T}
\label{eq:transformation}
\end{equation}

Following the same steps, 3D coordinates $(X',Y',Z')$ under the view of $v_{H,\theta}$ can also be transformed to 2D coordinates correspondingly, denoted as $(u',v')$. All of this, the ultimate goal is to build the relation between $(u,v)$ and $(u',v')$. After the view transformation, there is a relationship that has already been known, which can be written as:
\begin{equation}
\left\{ \begin{aligned}
X'&=X\\
Y'&=(Y-H)cos\theta+Zsin\theta\\
Z'&=-(Y-H)sin\theta+Zcos\theta
\end{aligned} \right.
\end{equation}
Besides, same as Equation~\ref{eq:transformation}, $(u',v')$ can be achieved by:
\begin{equation}
[u',v',1]^{T}=(Z')^{-1}K[X',Y',Z']^{T}
\label{eq:retransformation}
\end{equation}
From Equation~\ref{eq:transformation} to Equation~\ref{eq:retransformation}, the \textit{j-th} location $l_{j,f}$ with the coordinates $(u,v)$ on the front view map $I_F$ could be related to the corresponding location $l_{i,b}$ on the bird view map $I_B$, that is $(u',v')$. Besides, as discussed in Section~\ref{section:bird}, each location $l_{i,b}$ has been assigned with the probability score of pedestrian location and the corresponding width, described as $\left \{ p_{i, b}, w_{i, b} \right \}$, which could be radiated to $l_{j,f}$.

Above-mentioned formulations are designed to tackle the first task in "\textbf{\textit{What}}" branch: what the width of each location is on the front view map. There remains the other issue: what about the height is. Informally speaking, for the specific task of pedestrian detection, the difference in pedestrian height is, to a great extent, depth difference, in which the scale of an instance in the image is inversely proportional to the distance from the camera~\cite{ZhaoCrowd}. In another word, it means the farther away from the camera, the smaller the scale of the target will be. This fact allows us to model interdependency between depth and scale, and treat the process of proposal generation as an uneven distribution problem of scale.

First, still starting from Equation~\ref{eq:transformation}, the pedestrian height $\Delta h$ in one image could be modeled as the distance from head $\textbf{d}_1=[u_1, v_1]^T$ to foot $\textbf{d}_2=[u_2, v_2]^T$, formulated as $\Delta h=||\textbf{d}_1-\textbf{d}_2||_2$, while the real height $\Delta H$ in the real world could be mapped to $C\Delta H=Z\Delta h$, in which $C$ is a constant composed of camera intrinsic parameters and $Z$ is the depth we mentioned. As analyzed above, $Z$ (depth) and $\Delta h$ (height) are completely inverse when $C\Delta H$ is the fixed value. However, due to the influence of gender and age in pedestrian height, it is slightly unreasonable to set $\Delta H$ as a fixed value directly. In our height estimates, we assume that the height $\Delta H$ of all pedestrians obeys the uniform distribution and we analyze the error of the assumption. Previous studies of 63,000 European adults have shown that the average height of males and females is \textit{178cm} and \textit{165 cm} respectively, with a standard deviation of \textit{7cm} in both cases~\cite{VisscherSizing}. Besides, the distribution of human stature follows a Gaussian distribution for male and female populations~\cite{FreemanCross}, denoted as: $H_{male}\sim N(\mu_1 ,\sigma ^{2})$ and $H_{female}\sim N(\mu_2 ,\sigma ^{2})$, respectively. Specifically, for each point $l_{j,f}$ in the image $I_F$, its pedestrian height $\Delta H$ in real world is a sample from the above distribution, and we utilize the normal distribution function to evaluate the uncertainty of the sampled height $\Delta H$. Take the male distribution $H_{male}$ as an example. if the sampling height is $\hat{H}$, the corresponding uncertainty can be formulated as:
\begin{equation}
    \hat{e}=P(|x-\mu_{1}|<|\hat{H}-\mu_{1}|)=2\Phi (\frac{|\hat{H}-\mu_{1}|}{\sigma })-1
\label{e}
\end{equation}
in which $\Phi(\cdot)$ is function of standard normal distribution. The above $\hat{e}$ models the estimation uncertainty of $\Delta h$ due to variation of pedestrian height $\Delta H$ in real world, in which $\hat{e}$ goes smaller when $\hat{H}$ gets closer to $\mu_1$, and instead goes larger. This estimation uncertainty also servers as the confidence value to re-weight loss function, described in Section~\ref{loss}. With the above formulated $w$ and $h$, the corresponding proposal in the location of $(x,y)$ could be generated, and subsequently leads to the following "\textbf{\textit{Whether}}" branch.
\subsection{Corpus: Whether the target is a pedestrian}
After the "\textbf{\textit{Where}}" and "\textbf{\textit{What}}" branch, a large number of proposals have been generated, and the corresponding possibility score $p_{j,b}$, which indicates whether there is an pedestrian instance or not in the bird view map, could also be projected to the front view. However, the single bird view prediction has the problem: \textit{the increase of false positives}. Therefore, we introduce visual and attribute embedding, namely "\textbf{\textit{Whether}}" branch, to alleviate the above problem. This branch is based on two observations: one is that many false positives are mainly with human-like vertical structures, such as railings or tree trunks~\cite{Zhang2016How}; another is the fact that pedestrians, which perform walking or standing with up-straight pose, have one specific body structure from top to bottom: Head-Arm-Body-Leg. With this pedestrian-specific structure, we attempt to re-encode both occluded and non-occluded ones into the unified representation to distinguish those false positives with global vertical structure but lacking human-body based partial features.

The component of the proposed "\textbf{\textit{Whether}}" is shown in Figure~\ref{fig:framework}. It consists of a variational autoencoder (VAE)~\cite{Kingma2013VAE}, which is optimized by $L_{DE}$ to learn a lantent space shared by both visual features and attribute embeddings.
 More specifically, the embedding network takes the attributes embedding vector as input, and after passing through two fully connected (\textit{fc}) layers and the Rectified Linear Unit (\textit{ReLU}), outputs a visual embedding vector, which has the same dimensions as the visual feature vector of each proposal extracted from the encoder, optimizing:
\begin{equation}
    L_{DE}=\frac{1}{N}\sum _{i=1}^{N}||E(x_{i})-\varphi (y)||_{2}^{2}
\end{equation}
where $N$ is the total number of all proposals, $x_{i}$ is the visual feature of the i-th proposal, $y$ represents the word vector of pedestrian attributes including Head, Arm, Body and Leg, $E(\cdot)$ represents the visual feature encoder and $\varphi(\cdot)$ is the embedding network. The output feature via the decoder network used for the following classification and regression predicts the possibility score of whether there is a pedestrian or not, denoted as $p_{j, f}$, which is combined with $p_{j,b}$ and thus the overall possibility score of the j-th proposal could be formulated as:
\begin{equation}
    p_{j}=\frac{1}{2}(p_{j,b}+p_{j,f})
\end{equation}
\subsection{Training Details}
\label{loss}
Towards this task, we resort to depth maps derived from
a CNN-based monocular depth prediction model~\cite{abarghouei18monocular}. The feature generator consists of FPN~\cite{lin2016fpn} with a powerful backbone ResNet-50~\cite{He2016Resnet}, and the overall framework is optimized by the following loss function.

In the "\textbf{\textit{Where}}" branch, according to Equation~\ref{eq:transformation} to~\ref{eq:retransformation}, the horizontal and vertical central axes of each ground truth are mapped to the bird view, and thus constitute the bird view ground truth. Same as the loss function in CSP~\cite{liu2018CSP}, we also formulate possibility and width prediction as a classification and regression task via the cross-entropy and smooth L1 loss, respectively, denoted as $L_{bird}$ and $L_{width}$.

In the "\textbf{\textit{What}}" branch, it is actually a geometric transformation without extra \textit{conv} layers introduced. It is worth noting that in order to guarantee the robustness of the detector, pedestrian height $\Delta H$ in the real world is sampled from a distribution rather than using a fixed height value. Therefore, in order to evaluate the uncertainty of different samples, we introduce $\hat{e}$ in Equation~\ref{e} and consequently re-weight it on the following classification and regression loss.

In the "\textbf{\textit{Whether}}" branch, features extracted from each proposal are first encoded to a latent space, and subsequently decoded as the final classification and regression features, optimized by the loss $L_{cls}$ and $L_{reg}$. The overall loss can be formulated as:
\begin{equation}
\begin{aligned}
    L=&\lambda_1 L_{bird}+\lambda_2L_{width}+\lambda_3L_{DE}\\
    &+\frac{\lambda_4}{N}\sum _{i=1}^{N}(1-\hat{e}_{i})(L_{cls, i}+L_{reg,i})
\end{aligned}
\end{equation}
where $i$ represents the i-th proposal and $N$ is the total number of all proposals. $\lambda_1$, $\lambda_2$, $\lambda_3$ and $\lambda_4$ are the weights for each loss, which are experimentally set as 0.01, 0.1, 0.1 and 1, respectively.

\begin{table}[t]
\centering
\begin{tabular}{l|l|l|l}
\hline
$\theta$ & $Height$ & $Reasonable$ & $Heavy$\\
\hline
\multirow{2}*{$10\degree$} &5&21.3&63.4\\
\cline{2-4}
~ &20&20.8&60.8\\
\hline
\multirow{2}*{$30\degree$} &5&18.2&50.6\\
\cline{2-4}
~ &20&14.3&33.6\\
\hline
\multirow{2}*{$60\degree$} &5&15.5&34.9\\
\cline{2-4}
~ &20&{\color[rgb]{1,0,0}\textbf{9.3}}&{\color[rgb]{1,0,0}\textbf{18.7}}\\
\hline
\multirow{2}*{$90\degree$} &5&16.7&40.9\\
\cline{2-4}
~ &20&{\color[rgb]{0,0,1}\textbf{13.5}}&{\color[rgb]{0,0,1}\textbf{26.9}}\\
\hline
\end{tabular}
%}
\caption{Comparisons of the bird view $v_{\theta, H}$ on Citypersons, in which $\theta$ represents the depression angle and $Height$ represents the height of camera from horizontal plane. {\color[rgb]{1,0,0}\textbf{Boldface}}/{\color[rgb]{0,0,1}\textbf{Boldface}} indicate the {\color[rgb]{1,0,0}\textbf{best}}/{\color[rgb]{0,0,1}\textbf{second}} best performance.}
\label{comparetheta}
\end{table}

\begin{table}[t]
\centering
\begin{tabular}{l|l|l|l|l|l}
\hline
$w_{b}$&$w_{f}$ & $h_{b}$ & $h_{f}$ & $Reasonable$ &  $Heavy$\\
\hline
\checkmark&&&&{\color[rgb]{1,0,0}\textbf{9.3}}&{\color[rgb]{1,0,0}\textbf{18.7}}\\
&\checkmark&&&10.5&29.0\\
\checkmark&&\checkmark&&10.3&25.8\\
\checkmark&&&\checkmark&{\color[rgb]{0,0,1}\textbf{9.8}}&{\color[rgb]{0,0,1}\textbf{20.1}}\\
&\checkmark&\checkmark&&12.3&43.4\\
&\checkmark&&\checkmark&11.0&38.6\\
\hline
\end{tabular}
%}
\caption{Comparisons of different prediction combinations of $(w_b,h_b)$ and $(w_f,h_f)$ on Citypersons, which represents $(width,height)$ of bird view and front view, respectively. {\color[rgb]{1,0,0}\textbf{Boldface}}/{\color[rgb]{0,0,1}\textbf{Boldface}} indicate the {\color[rgb]{1,0,0}\textbf{best}}/{\color[rgb]{0,0,1}\textbf{second}} best performance.}
\label{comparewhbf}
\end{table}

\section{Experiments}
We assess the effectiveness of our proposed method for pedestrian detection on widely used datasets Cityperson~\cite{Zhang2017CityPersons} and Caltech~\cite{Dollar2009Pedestrian}. Results are the MR$^{-2}$ evaluation metric, in which \textbf{lower is better}.
\subsection{Experiment Setup}
\textbf{Datasets} Citypersons ~\cite{Zhang2017CityPersons} is a diverse dataset built
upon the Cityscapes data, which includes 5000 images (2975 for training, 500 for validation, and 1525 for testing). In a total of 5 000 images, it has $\sim$35k person and $\sim$13k ignore region annotations. And it notices the density of persons are consistent across
train/validation/test subsets. The Caltech Dataset ~\cite{Dollar2009Pedestrian} consists of approximately 10 hours of 640x480 30Hz video taken from a vehicle driving through regular traffic in an urban environment. About 250,000 frames with a total of 350,000 bounding boxes and 2300 unique pedestrians were annotated. All of the datasets contain challenging settings, denoted as Heavy occlusion, in which the propotion of visible parts of pedestrians is less than $0.65$.

\textbf{Implementation} We implemented the proposed method in Pytorch with the backbone ResNet50~\cite{He2016Resnet} and the Nvidia GTX1080Ti. We optimize the network using the Stochastic Gradient Descent (SGD) algorithm with 0.9 momentum and 0.0005 weight decay, respectively. For Citypersons, the mini-batch contains 2 images and we train the network for $30k$ iterations with the initial learning rate of $10^{-3}$ and decay it to $10^{-4}$ for another $6k$ iterations. For Caltech, the mini-batch contains 8 images and we train the network for $40k$ iterations with the initial learning rate of $10^{-3}$ and decay it to $10^{-4}$ for another $20k$ iterations.

\subsection{Ablation}
In this section, we evaluate how each significant component
of our network contributes to performance using the Citypersons dataset under the Reasonable and Heavy settings, which are specification for the non-occluded and occluded pedestrian. It is worth noting that to evaluate the performance of each branch, such as results in Table~\ref{comparewww-www}, we take each "\textbf{\textit{W}}" step as one enhanced component, and replace the corresponding module in baseline (Faster RCNN, Table~\ref{comparewww-www} Line~1). For example, to get the result of "\textbf{\textit{Where}}", the process of proposal generation in RPN is replaced by 'head' and width prediction in the bird view map together with fixed aspect ratio. Or to evaluate the "\textbf{\textit{What}}" step, we replace the fixed aspect ratio with the depth based scale estimation. The same is true for experiments in Table~\ref{comparewww-csp}.

\begin{table}[t]
\centering
\begin{tabular}{l|l|l|l|l}
\hline
$Where$ & $What$ & $Whether$ & $Reasonable$ &  $Heavy$\\
\hline
&&&14.6&60.6\\
\checkmark & & & 11.0 & 23.3\\
&\checkmark&&12.4&52.6\\
&&\checkmark& 10.8 & 30.2\\
\checkmark&\checkmark&& 9.9 & 21.7\\
\checkmark&&\checkmark& 10.3 & 20.5 \\
\checkmark&\checkmark&\checkmark&9.3&18.7\\
\hline
\end{tabular}
%}
\caption{Ablation study of W$^3$Net on Citypersons.}
\label{comparewww-www}
\end{table}

\begin{table}[t]
\centering
\begin{tabular}{l|l|l|l|l}
\hline
$Where$ & $What$ & $Whether$ & $Reasonable$ &  $Heavy$\\
\hline
\checkmark & & & 10.9 & 24.6\\
&\checkmark&& 10.3 & 45.2 \\
&&\checkmark& 11.0 & 31.0 \\
\hline
\end{tabular}
%}
\caption{Ablation study of CSP on Citypersons.}
\label{comparewww-csp}
\end{table}

\begin{table}[t]
\centering
\begin{tabular}{l|l|l}
\hline
$Method$ & $Reasonable$ &  $Heavy$\\
\hline
W$^3$Net (with $G_{A\rightarrow C}$)&13.4&32.1\\
W$^3$Net (with $G_{A\rightarrow B }$ and $G_{B\rightarrow C}$)&9.3&18.7\\
\hline
\end{tabular}
%}
\caption{Ablation study of bird view generation with or without domain transfer from real-world to synthetic data on Citypersons.}
\label{compareGAN}
\end{table}

\textbf{Why is the bird view?} Occlusion is one of the great challenges entrenched in pedestrian detection, which is ultimately caused by isolated view point, especially the front view. However, we could actually discover that in the real 3D world, even if a pedestrian is occluded from the initial view, bird view still performs free from occlusion issues. Besides, pedestrian instances are condensed to a point on the bird view, namely "head", which can be seamlessly connected with the anchor-free detectors. Inspired by above, experiments are conducted and comparisons are reported in Table~\ref{comparewww-www}. The method combined with "\textbf{\textit{Where}}" branch achieves 23.3$\%MR^{-2}$ on Heavy occlusion subset, which is an absolute 37.3-point improvement over our baseline 60.6\%. When the "\textbf{\textit{Where}}" branch is added to CSP, as is shown in Table~\ref{comparewww-csp}, the performance of CSP is boosted by a large margin from 49.3$\%MR^{-2}$ to 24.6$\%MR^{-2}$. It is in accordance with our intuition that "\textbf{\textit{Where}}" branch is specially designed for the long-standing occlusion problem.
In addition, ablations are carried out in Table~\ref{comparetheta} and~\ref{compareGAN} to investigate whether the difference in generation of bird view map will influence the results. Comparisons reported in Table~\ref{comparetheta} show that the view $v_{\theta, H}$ under $\theta=60\degree$ and $H=20$ meters achieves the best performance, and we believe that more detailed exploration of the parameters $\theta$ and $H$ can further improve the performance, but it is not in the scope of this work. In Table~\ref{compareGAN}, the method without domain transfer directly generate bird view images from the real-world data suffering a great decline on both Reasonable and Heavy subsets, which demonstrates the effectiveness of the proposed $G_{A\rightarrow B}$.

\begin{figure}
\begin{center}
\includegraphics[width=8.5cm]{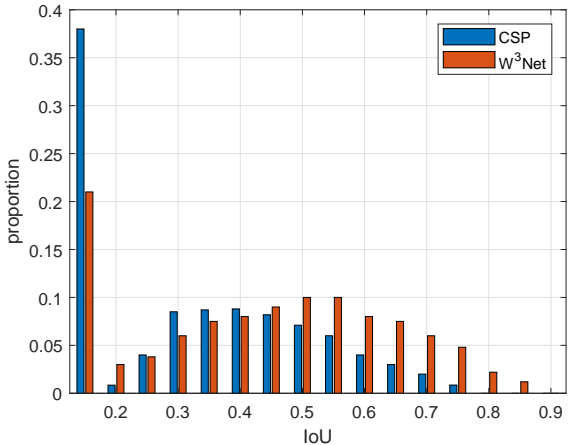}
\end{center}
   \caption{IoU histogram of different methods, CSP and the proposed W$^3$Net, in which shows the W$^3$Net generates more high quality proposals with $IoU>0.5$.}
\label{fig:iou}
\end{figure}

\begin{table}[t]
\centering
\begin{tabular}{l|l|l|l}
\hline
& $Reasonable$ & $Heavy$&$Test\ Time$\\
\hline
TLL~\cite{TLL2018}&15.5&53.6&-\\
FRCNN~\cite{Zhang2018Attention}&15.4&-&-\\
TLL+MRF~\cite{TLL2018}&14.4&52.0&-\\
RepLoss~\cite{Wang2017Repulsion}&13.2&56.9&-\\
OR-CNN~\cite{Zhang2018Occlusion}&12.8&55.7&-\\
ALFNet~\cite{Liu2018ALF}&12.0&51.9&0.27s/img\\
CSP~\cite{liu2018CSP}&{\color[rgb]{0,0,1}\textbf{11.0}}&{\color[rgb]{0,0,1}\textbf{49.3}}&0.33s/img\\
\hline
W$^3$Net(ours)&{\color[rgb]{1,0,0}\textbf{9.3}}&{\color[rgb]{1,0,0}\textbf{18.7}}&0.31s/img\\
\hline
\end{tabular}
%}
\caption{Comparisons with other state-of-the-art methods on Citypersons. {\color[rgb]{1,0,0}\textbf{Boldface}}/{\color[rgb]{0,0,1}\textbf{Boldface}} indicate the {\color[rgb]{1,0,0}\textbf{best}}/{\color[rgb]{0,0,1}\textbf{second}} best performance.}
\label{compareCity}
\end{table}

\textbf{Why is the depth?} Depth is a useful auxiliary information, however, it has not been fully identified and utilized in pedestrian detection. One is that depth implies scale cues, especially for pedestrians, which share strong intra-class similarity. The other is that depth plays a connecting role in our framework. Depth, that is "\textbf{\textit{What}}" branch, projects detection results from "\textbf{\textit{Where}}" branch to the front view, and on the basis of interdependency between depth and scale, generating reasonable proposals for the following "\textbf{\textit{Whether}}" branch. As shown in Table~\ref{comparewww-www} and~\ref{comparewww-csp}, compared with the baseline or the bare CSP, the proposed "\textbf{\textit{What}}" branch goes a step further and achieves superior performance on both subsets. It is still worth noting that "\textbf{\textit{What}}" branch is just a kind of matrix transformation without any \textit{conv} or \textit{fc} layers introduced, which preserves the effectiveness of the detector while ensuring better performance. To further investigate the effect, we also conduct experiments including a prediction network to estimate pedestrian height, as reported in Table~\ref{comparewhbf}, in which the single $w_b$ means the detector combines width prediction on bird view with height estimation by depth-scale relations, while $w_b+h_f$ combines both width and height prediction on bird view and front view, respectively. It can be observed that other methods can achieve comparable but suboptimal results to $w_b$ prediction. This result may be attributed to the accurate and consistent height estimation with less noise during training, which has also been proved in Figure~\ref{fig:iou}. Proposals generated by the proposed method shares better \textit{iou} performance, in which the proportion of $IoU>0.5$ is improved greatly.

\begin{figure*}
\begin{center}
\includegraphics[width=17cm]{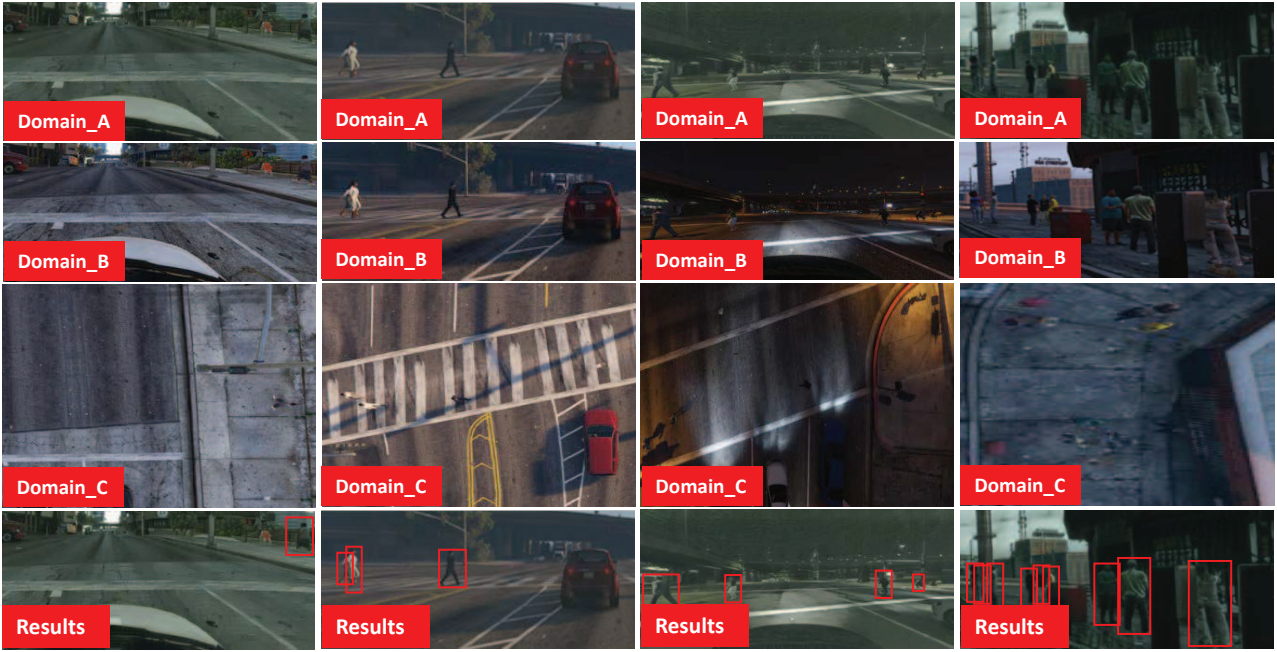}
\end{center}
   \caption{Visualization results of the proposed method, in which Domain\_A, Domain\_B and Domain\_C represent the front view map, the domain-transformed map and the bird view map, respectively. Red bounding boxes represent the overall detection results of the W$^3$Net.}
\end{figure*}

\textbf{Why is the corpus?} A natural question is, for occlusion, is the bird view enough? The bird view map can effectively provide the locations of all pedestrians. However, due to ambiguous pedestrian features, suspected objects, such as railings, which are similar to pedestrian 'head' in the bird view, could also be falsely classified as the pedestrian. In order to tackle this problem, we introduce the "\textit{\textbf{Whether}}" branch, which falls into the proposal detection fashion on the basis of the front view map. The difficulty lies in proposal classification is that features of occluded pedestrians usually perform incomplete compared with those of non-occluded instances, that is to say, these two types have different feature distribution. Therefore, we re-encode proposals into a unified corpus space, which with fixed dimension and clear semantic information generates robust relations between visual and semantic cues and thus benefits robust feature representation. Results are reported in Table~\ref{comparewww-www} and~\ref{comparewww-csp}. It can be seen that although "\textbf{\textit{What}}" and "\textbf{\textit{Whether}}" branch both focus on occluded pedestrian detection, their detection uses different modal input, which maximises the complementary information of different views and thus achieve the best performance finally.
\begin{table}
\centering
\begin{tabular}{l|l|l|l|l}
\hline
& $R^{O}$ & $H^{O}$ & $F^{O}$ & $R^{N}$\\
\hline
ACF++~\cite{OhnBar2016Boost}&17.7&79.51&100&14.68\\
DeepParts~\cite{felzenszwalb2010dpm}& 11.89 &60.42 & 100 & 12.9\\
FasterRCNN+ATT~\cite{Zhang2018Attention}& 10.33 & {\color[rgb]{0,0,1}\textbf{45.18}}  & 90.94 & 8.11\\
MS-CNN~\cite{Cai2016Multiscale}& 9.95 &59.94 & 97.23 & 8.08\\
RPN+BF~\cite{Zhang2016RPNBF}& 9.58 &74.36 & 100 & 7.28\\
TLL~\cite{TLL2018}& 8.45 & - &  {\color[rgb]{0,0,1}\textbf{68.03}}  & -\\
SDS-RCNN~\cite{SDSRCNN2017}&  {\color[rgb]{0,0,1}\textbf{7.36}} & 58.55 & 100 & 6.44\\
RepLoss~\cite{Wang2017Repulsion}&-&-&-&{\color[rgb]{0,0,1}\textbf{4.0}}\\
OR-CNN~\cite{Zhang2018Occlusion}&-&-&-&4.1\\
\hline
W$^3$Net(Ours)&{\color[rgb]{1,0,0}\textbf{6.37}}& {\color[rgb]{1,0,0}\textbf{28.33}} & {\color[rgb]{1,0,0}\textbf{51.05}} & {\color[rgb]{1,0,0}\textbf{3.82}} \\
\hline
\end{tabular}
\caption{Comparisons with the state-of-the-art methods on Caltech dataset. $*^{O}$ means the result is under the standard(old) test annotations, and $*^{N}$ means the result is under the new annotations provided by~\cite{Zhang2016How}. $R$, $H$, and $F$ represent the subset of $Reasonable$, $Heavy$, $Far$ targeting at non-occluded, occluded and small-scale pedestrian detection, respectively. {\color[rgb]{1,0,0}\textbf{Boldface}}/{\color[rgb]{0,0,1}\textbf{Boldface}} indicate the {\color[rgb]{1,0,0}\textbf{best}}/{\color[rgb]{0,0,1}\textbf{second}} best performance.}
\label{Caltech}
\end{table}
\subsection{Comparisons with the State-of-the-arts}
Performance compared with state-of-the-art methods on Citypersons Reasonable, Partial and Heavy dataset is shown in Table~\ref{compareCity}, while the result on Caltech is reported in Table~\ref{Caltech}. It can be observed that \emph{1)} W$^3$Net leads a new state-of-the-art result of 9.3\% (Citypersons) and 3.82\% (Caltech) on both Reasonable subsets.It is note worthy that at the same time, the proposed method also achieves promising performance on occluded instances. In particular,when evaluating heavy occlusion subset on Citypersons, our model produces18.7\% while prior arts range from 49\%-56\%. \emph{2)} Result on Far subset of Caltech outperforms previous advanced detectors, such as TLL~\cite{TLL2018} by 16.98$\%$ which is specially designed for small-scale targets, demonstrates the superiority on scale variation. \emph{3)} Despite introducing multi-modal data, the careful discovery of depth relation actually reduces computational cost without extra \textit{conv} or \textit{fc} layers, and thus the inference time on one GTX 1080Ti with the 1x. image scale is still in line with the state-of-the-arts.

\section{Conclusion}
In this paper, we propose a novel network confronts challenges (occlusion and scale variation) preoccupied in pedestrian detection, which decouples the task into Where, What and Whether problem directing against pedestrian localization, scale estimation  and classification correspondingly. As a result, the detector achieves the new state-of-the-art performance under various challenging settings. The framework is a customized design for pedestrian detection, but it is easy to extend into other tasks like face or vehicle detection, which need re-model interdependency between depth and scale, and is already in our future plans.

\section{Acknowledgments}
This work was partially funded by the National Science Fund of China under Grant No.61971281, the National Key Research and Development Program No.2017YFB1002401, STCSM No.18DZ1112300 and No.18DZ2270700.
{\small
\bibliographystyle{ieee_fullname}
\bibliography{egbib}
}

\end{document}